\PassOptionsToPackage{table}{xcolor}
\PassOptionsToPackage{colorlinks=true, linkcolor=blue, citecolor=blue, urlcolor=blue}{hyperref}

\documentclass[sigplan,twocolumn,nonacm]{acmart}

\renewcommand\footnotetextcopyrightpermission[1]{}
\setcopyright{none}
\acmConference[EuroSys '27]{The 22nd European Conference on Computer
Systems}{2027}{Rabat, Morocco}

\usepackage{amsmath, amssymb, amsfonts}
\usepackage{bm}
\usepackage{booktabs}
\usepackage{multirow}
\usepackage{tabularx}
\usepackage{colortbl}
\usepackage{graphicx}
\usepackage{placeins}
\usepackage{xurl}
\usepackage{xspace}
\usepackage{microtype}
\usepackage{tikz}
\usetikzlibrary{shapes, arrows.meta, positioning, calc, decorations.pathreplacing, patterns}
\usepackage{pgfplots}
\pgfplotsset{compat=1.17}
\AtBeginDocument{%
  \setlength{\abovedisplayskip}{3.5pt plus 1.5pt minus 1.5pt}%
  \setlength{\belowdisplayskip}{3.5pt plus 1.5pt minus 1.5pt}%
  \setlength{\abovedisplayshortskip}{2pt plus 1pt}%
  \setlength{\belowdisplayshortskip}{2.5pt plus 1pt minus 1pt}%
}

\usepackage{enumitem}
\setlist{topsep=2pt, itemsep=1pt, parsep=0pt, partopsep=0pt, leftmargin=*}

\usepackage{etoolbox}
\makeatletter
\patchcmd{\@afterheading}{\clubpenalty \@M}{\clubpenalty 2500}{}{%
  \PackageWarning{preamble}{could not relax the post-heading club penalty}}
\makeatother

\newcommand{\code}[1]{\path{#1}}

\DeclareRobustCommand{\methodname}{\mbox{DuplexCadence}\xspace}

\newcommand{\regions}{\mathcal{R}}

\newcommand{\bytes}{\operatorname{bytes}}

\newcommand{\catalog}{\mathcal{K}}

\definecolor{cellgray}{gray}{0.93}
\definecolor{sbblue}{RGB}{41, 114, 196}
\definecolor{sbred}{RGB}{226, 111, 82}
\definecolor{sbgreen}{RGB}{111, 168, 104}

\makeatletter
\renewcommand{\@mkauthors}{%
  \global\setbox\mktitle@bx=\vbox{%
    \noindent\unvbox\mktitle@bx\par\medskip\centering
    \normalsize\bfseries
    Haixiao Gao\textsuperscript{1},
    Yimin Zheng\textsuperscript{1},
    Linyou Xiao\textsuperscript{1},
    Zeke Xie\textsuperscript{2*}\par
    \vspace{0.35em}%
    \normalsize\normalfont
    \textsuperscript{1}Jinan University, China\par
    \vspace{0.15em}%
    \textsuperscript{2}Hong Kong University of Science and Technology (Guangzhou), China\par
    \vspace{0.35em}%
    \{ghx2004, zymmm, pingfkl\}@stu2023.jnu.edu.cn,
    zekexie16@gmail.com\par
    \bigskip}}
\makeatother

\AtBeginDocument{% Supplement cross-reference numbers. The supplement itself is not
\newlabel{eq:admit}{{S3}{13}{The admission predicate}{equation.G.3}{}}
\newlabel{eq:representation}{{S2}{13}{Execution model and representation invariant}{equation.G.2}{}}
\newlabel{fig:lifecycle}{{S2}{11}{\textbf {The host-side path of one chunk} (\S \ref {sec:design:replay}). Both rules on the path the host walks per emitted chunk. The expensive step, staging the stacked carry, disappears at the \texttt {aliased} line, and only because the state rule owns that storage. Clauses (i)--(iii) are the three conjuncts of \eqref {eq:admit}, evaluated in order and fail-closed; every \texttt {eager} carries a named reason and is counted (\S \ref {sec:design:verify}). In steady state none of the three eager branches is taken and a chunk is exactly three replays, one per wrapped callable, with only the \texttt {istft} tail left eager. The RNG probe, side-stream warmup, and shared graph pool inside \texttt {capture} are described in \S \ref {sec:design:replay}.\relax }{figure.caption.15}{}}
\newlabel{fig:pareto}{{S1}{10}{\textbf {\methodname is the only configuration that is simultaneously fastest, memory-saving, and byte-exact.} The Pareto view of Table~\ref {tab:factorial}. Each remedy alone moves along one axis, and the axis of shrinking state to demand is memory: it reaches a lower peak than \methodname but buys no speed. Only the coupling (solid arrow) moves along both. Token2Wav decoder, A100. The step-clock cells R and SR are omitted for legibility; both sit in the same vertical band at 86.3\,ms and neither changes the ordering. Compiler and vendor points are medians of the $n{=}4$ matched control sweep of Table~\ref {tab:controls}, whose same-sweep B0/SRc pair is 245.3/84.5\,ms and preserves the ordering.\relax }{figure.caption.13}{}}
\newlabel{sec:appendix:admission}{{S2.6}{4}{Replay admission statistics}{subsection.B.6}{}}
\newlabel{sec:appendix:attribution}{{S3.2}{6}{Per-mechanism attribution of the segment speedup}{subsection.C.2}{}}
\newlabel{sec:appendix:bounds}{{S6.1}{10}{Bound arithmetic for Table~\ref {tab:slack}}{subsection.F.1}{}}
\newlabel{sec:appendix:capacity}{{S8}{14}{Shared-Weight Serving Capacity}{appendix.H}{}}
\newlabel{sec:appendix:carry}{{S2.7}{4}{Representation-invariant verification}{subsection.B.7}{}}
\newlabel{sec:appendix:controlci}{{S6.5}{12}{Paired intervals for the control sweep}{subsection.F.5}{}}
\newlabel{sec:appendix:determinism}{{S2.9}{5}{Determinism protocol, and when a gate is vacuous}{subsection.B.9}{}}
\newlabel{sec:appendix:displaced}{{S6}{10}{Displaced Figures, Tables, and Derivations}{appendix.F}{}}
\newlabel{sec:appendix:enabling}{{S3.1}{5}{Pricing the Enabling Penalty from the Declaration}{subsection.C.1}{}}
\newlabel{sec:appendix:enablingderiv}{{S7.3}{13}{Growth of the staging term with granularity}{subsection.G.3}{}}
\newlabel{sec:appendix:engine}{{S5}{7}{Implementation Details}{appendix.E}{}}
\newlabel{sec:appendix:families}{{S7.6}{14}{The carry on the other two families}{subsection.G.6}{}}
\newlabel{sec:appendix:freezeomni}{{S3.6}{7}{Freeze-Omni Transfer Detail}{subsection.C.6}{}}
\newlabel{sec:appendix:hazards}{{S5.6}{9}{Hazards the contract absorbs}{subsection.E.6}{}}
\newlabel{sec:appendix:invariant}{{S7.1}{12}{Execution model and representation invariant}{subsection.G.1}{}}
\newlabel{sec:appendix:l20}{{S4}{7}{L20 Detail Tables}{appendix.D}{}}
\newlabel{sec:appendix:limits}{{S6.6}{12}{Limitations, in detail}{subsection.F.6}{}}
\newlabel{sec:appendix:membership}{{S7.5}{14}{Class membership of the released stacks}{subsection.G.5}{}}
\newlabel{sec:appendix:moshi}{{S3.5}{7}{Moshi Per-Seed Replay Toggle}{subsection.C.5}{}}
\newlabel{sec:appendix:orule}{{S6.3}{11}{The optional schedule rule}{subsection.F.3}{}}
\newlabel{sec:appendix:overreserve}{{S6.2}{11}{Per-row derivations for Table~\ref {tab:overreserve}}{subsection.F.2}{}}
\newlabel{sec:appendix:porting}{{S5.1}{7}{Porting cost}{subsection.E.1}{}}
\newlabel{sec:appendix:proofs}{{S1}{1}{Proofs}{appendix.A}{}}
\newlabel{sec:appendix:protocol}{{S2}{3}{Artifact and Protocol Details}{appendix.B}{}}
\newlabel{sec:appendix:provenance}{{S6.4}{11}{Measurement provenance for the shared decoder}{subsection.F.4}{}}
\newlabel{sec:appendix:residency}{{S5.7}{9}{Device-residency normalization}{subsection.E.7}{}}
\newlabel{sec:appendix:rng}{{S2.8}{5}{Replay and the device generator}{subsection.B.8}{}}
\newlabel{sec:appendix:scope}{{S7.4}{13}{Scope conditions and the excluded rows}{subsection.G.4}{}}
\newlabel{sec:appendix:sessions}{{S8.2}{15}{Multi-session bindings, in full}{subsection.H.2}{}}
\newlabel{sec:appendix:setup}{{S2.2}{3}{Full experimental setup}{subsection.B.2}{}}
\newlabel{tab:adapter}{{S11}{16}{\textbf {What one adapter declares} (\S \ref {sec:design:adapter}). The complete declaration for the shared streaming decoder at the operating point of \S \ref {sec:eval:setup}; clock quantities are in frames, the upsampling rate in frames per token. $f_{\text {chunk}}$ is the retained advance per emitted chunk (25 tokens) and $f_{\text {call}}$ the largest extent one call presents (25 tokens plus a 3-token lookahead); the pair is what puts live traffic outside the stock bucket set. The two address-stable regions are the carry and the solver workspace; the three device-resident constants are the sinusoid frequency table, the STFT window, and the harmonic multiplier.\relax }{table.caption.26}{}}
\newlabel{tab:capacity}{{S10}{14}{\textbf {Intra-session savings convert into serving capacity.} A100 shared-weight harness; G6 PASS and 100\% paired PCM equality on every completed cell. Peak gap widens with $N$ under shared weights. $^{\dagger }$SRc completes all six overloaded blocks (feasibility). $^{\ddagger }$p95 omitted where both arms miss $\approx $90\% of rounds.\relax }{table.caption.25}{}}
\newlabel{tab:certificate}{{S9}{12}{\textbf {Each premise of Proposition~\ref {prop:replay}, and the representation invariant the state rule rests on, is discharged by a named mechanism that reports itself} (\S \ref {sec:design:verify}). Binding premises are re-checked on every call and effect premises are settled once per class at capture; the representation invariant is shadowed against the stock path at every retention application (\S \ref {sec:appendix:carry}), and the determinism envelope is the model's own protocol, pinned identically on both arms wherever a comparison crosses processes. Every clause fails closed, so a violation costs one eager call at the active state layout, and the run's own counters say which clause fired.\relax }{table.caption.16}{}}
\newlabel{tab:o}{{S3}{7}{\textbf {Schedule-rule paired test} (A100; SRc vs.\ SRcO, the same variant with the optional schedule rule enabled; 8 matched blocks). Exactness and audio-ready gates pass; the wall-clock CI includes 0. Decision: optional.\relax }{table.caption.6}{}}
\newlabel{tab:phasepack}{{S2}{6}{\textbf {Composition with a phase-packing scheduler.} Mixed $N{=}2$ on A100 under an outer two-pass schedule, same block seeds as the serial shared-weight harness; 6/6 completed blocks. The reported delta is the median of six paired block differences, which is why it differs from the difference of the two arm medians.\relax }{table.caption.4}{}}
\newlabel{tab:protocol}{{S1}{6}{\textbf {The protocol changes the ratio but not the equality.} Shared decoder at the operating point of \S \ref {sec:eval:setup}, 20 measured chunks. Each row pins a different kernel protocol, so the three rows hash differently from \emph {each other}; within every row the eager and replay arms agree bitwise. Pinning costs the eager arm far more than the replay arm, because deterministic-algorithm checks are a host-side cost that a replayed graph does not pay, so the speedup measured under a pinned protocol is the larger number and \S \ref {sec:eval:factorial} reports the smaller one.\relax }{table.caption.3}{}}
}

\begin{document}

% acmart installs its theorem environments at the end of the preamble, so
% these have to follow \begin{document}.  The criterion environment shares
% the theorem counter but names the scope result for what it is: arithmetic
% on declared quantities, meant to be evaluated on a configuration file
% rather than read as a structural result.
\theoremstyle{acmplain}
\newtheorem{criterion}[theorem]{Design criterion}

\title[\methodname\ for Interactive Speech]{%
\methodname: Exact State and Execution from a Speech Model's Declared
Timelines}

\author{Haixiao Gao}
\email{ghx2004@stu2023.jnu.edu.cn}
\author{Yimin Zheng}
\email{zymmm@stu2023.jnu.edu.cn}
\author{Linyou Xiao}
\email{pingfkl@stu2023.jnu.edu.cn}
\affiliation{%
  \institution{Jinan University}
  \country{China}}
\author{Zeke Xie}
\email{zekexie16@gmail.com}
\affiliation{%
  \institution{Hong Kong University of Science and Technology (Guangzhou)}
  \country{China}}
\renewcommand{\shortauthors}{Gao et al.}

\begin{abstract}
Full-duplex speech models support streaming interaction that listens
and speaks at the same time.  Serving them is governed by a strict,
repeating deadline: conversation advances on a one-second cadence, and
every second of input must be turned into a second of speech before the
next second arrives.  Because stages within a session run in strict
sequence, per-invocation overhead cannot be batched away.  Profiling
reveals that the autoregressive stages of a duplex second already fit
within the period, whereas the token-to-audio synthesis tail is what
causes overruns.  This tail stage suffers from \emph{orchestration
slack} where the GPU is left waiting as thousands of tiny, regular
operations are issued one by one, while also wasting substantial
memory by over-provisioning state at static implementation constants.
Existing remedies, such as graph recording and demand-sized allocation,
fail because streaming state dynamics violate their prerequisites.
The root cause is that the runtime lacks the model's \emph{native
clocks}: the per-region counters that govern advancement rates and
retention policies.  We propose \methodname, which explicitly
declares native clocks to the runtime and derives two mutually enabling
rules: demand-sized state allocation at a stable address,
and exact-shape graph replay without padding.  The former eliminates
idle memory and stabilizes tensor pointers, while the latter removes
orchestration slack without padding overhead.  Evaluated on four
released models across three decoder architectures with bit-for-bit
identical output, \methodname reaches $\bm{2.85\times}$ the stock
runtime's speed at $\bm{38.8\%}$ lower peak memory.  On the live duplex
path, mean SPEAK time falls from 14\% over the one-second cadence to 2\%
under it, enabling models to reliably keep up with interactive speech
while markedly expanding multi-session serving capacity.
\end{abstract}

% acmart requires CCS concepts and keywords for papers over two pages.  The
% \ccsdesc lines below carry the classification; the machine-readable CCSXML
% block that ACM wants alongside them must be regenerated for the
% camera-ready from https://dl.acm.org/ccs, which emits the concept IDs.


\maketitle

\section{Introduction}
\label{sec:introduction}

Speech is an increasingly important interface between humans and machines.
From conversation with large foundation models to embodied robotics, a growing number of systems adopt speech as an interaction modality~\cite{cui2026minicpmo45,xu2025qwen25omni,wu2025stepaudio2}.
The current generation of speech models is full-duplex~\cite{defossez2024moshi,wang2025freezeomni,roy2026personaplex}: it listens and speaks simultaneously, allowing users to interrupt at any moment.
Released systems such as MiniCPM-o~4.5~\cite{cui2026minicpmo45}, Moshi~\cite{defossez2024moshi}, and Lychee-FD~\cite{liu2026lycheefd} are moving from laboratory demonstrations to deployed products.

To achieve real-time interaction, a speech system must begin speaking while it is still listening.
This shifts the serving objective from maximizing average throughput to meeting a periodic deadline: the conversation advances on a one-second cadence, and every second of input must be turned into a second of speech before the next second arrives.
Whether this deadline can be honored depends on how much work one second requires and whether that work can be amortized.
However, inside a single session, the execution stages exhibit strict sequential dependencies, with each stage consuming the output of the preceding one.
The standard mechanism in large-model serving for amortizing fixed per-invocation launch overhead is to batch independent requests together~\cite{yu2022orca,kwon2023vllm,zheng2024sglang,agrawal2024sarathi}.
Inside an individual session, no batching target exists.
The launch overhead lands in full on the critical path of the single session, making the cost of one second a quantity that must be audited stage by stage.

\begin{figure}[t]
\centering
\includegraphics[width=\columnwidth]{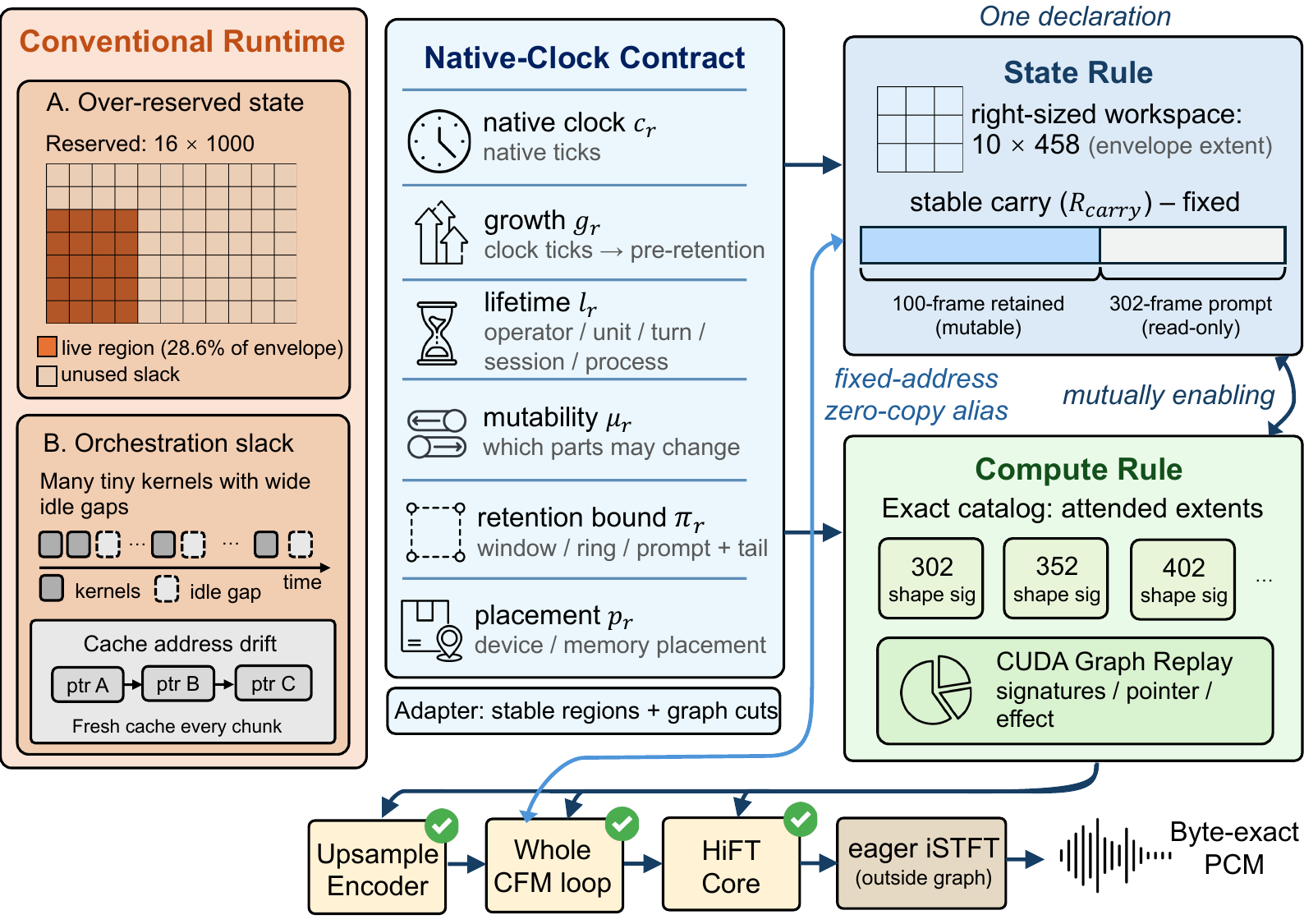}
\caption{\textbf{One second of full-duplex speech synthesis and recurring operations.}
Conventional runtimes over-provision state at static constants and issue operations individually.
\methodname declares native clock rates and retention bounds to derive demand-sized state at a stable address and an exact-shape replay plan without padding.}
\Description{Side-by-side motivation diagram.  The conventional runtime over-provisions a fixed state envelope, changes cache addresses, and issues many small GPU kernels.  DuplexCadence uses a native-clock contract to derive compact stable state and exact-shape graph replay, producing identical PCM bytes with lower latency and memory.}
\label{fig:motivation}
\end{figure}

We profile that critical path (\S\ref{sec:motivation}).
The autoregressive language modeling stages already keep up with the one-second cadence.
What pulls the entire stream past the deadline is the synthesis tail that converts speech tokens into audio waveforms; specifically, it spends about $1.17$ seconds on one second of work.
An analysis of this tail reveals that the system is not compute-bound, but rather bottlenecked by a starved accelerator: thousands of tiny, regularly shaped operations are handed to the GPU one by one, leaving it waiting for 57--65\% of wall-clock time.
We call this idle time \emph{orchestration slack}; it is what pushes a segment that would otherwise finish on time past the deadline.
The same execution exhibits a second severe resource waste in memory: the flow decoder actively uses only one-seventh of the workspace it allocates, holding the remainder occupied yet unused.
These two defects impose distinct constraints on full-duplex serving.
The former dictates whether an individual stream can meet the cadence, whereas the latter caps the number of concurrent sessions a device can sustain.
Figure~\ref{fig:motivation} contrasts both inefficiencies against what becomes possible once the model's rates are declared.

Each problem has an established general remedy.
To eliminate host-side launch gaps, systems commonly rely on graph recording (such as CUDA Graphs) to capture a recurring sequence of calls once and issue it as a whole~\cite{kwon2020nimble,ma2020rammer}.
To curb excessive memory envelopes, systems use demand-sized allocation to reserve storage only for the state kept live under a retention policy.
Each remedy relies on a strict precondition: graph recording requires invoked operations to exhibit fixed tensor shapes and stable pointer addresses, while demand-sized allocation requires the runtime to know which history remains live.
Neither precondition holds in streaming speech.
Persistent history state updates with every processed chunk: its length initially grows with newly arrived frames and is subsequently truncated once it exceeds the retention bound.
As a result, hot operators observe input lengths that vary across chunks, so the fixed shapes required by graph recording do not exist.
Existing implementations also reallocate memory and copy state on every chunk, causing physical addresses to drift dynamically and rendering recorded execution plans invalid.
Meanwhile, the runtime lacks visibility into the true liveness boundary defined by the retention policy, holding only static capacity constants hardcoded in the implementation.
Consequently, demand-sized allocation cannot be applied.
The stock CUDA Graph fast path provided in released runtimes never executes for precisely these reasons, silently falling back to slow, individual operator launches (\S\ref{sec:motivation}).

In hindsight, neither inefficiency should exist, because the underlying information has been available all along.
A single second of streaming synthesis advances deterministically along multiple parallel timelines: the language model steps once per token, the speech decoder steps once per audio frame, and waveform generation steps once per block of samples.
The rate of each timeline is fixed in the model's static configuration files, and the same configuration explicitly states the retention policy of every persistent state region, such as the number of retained history frames and prompt frames.
Because these rates are constant, specific small operations recur thousands of times per second with identical, fully enumerable dimensions.
What the runtime lacks is neither raw compute power nor better operator kernels, but rather these timelines themselves, specifically an explicit mechanism to expose the model's inherent temporal structure so that the runtime can act upon it.

\paragraph{Key insight.}
We propose \methodname, a runtime system that lifts these inherent timelines into explicit declarations that the runtime can directly interpret.
We term these declarations the model's \textbf{native clocks}: one deterministic counter per persistent state region, governing both the offset where new state is written next and the maximum extent a consumer operator may read backward.
A region's native clock, coupled with its configured retention policy, answers the two questions that previously blocked existing remedies: how large the region ever needs to be, and which specific slice shapes a consumer will ever access.
To resolve the first question, \methodname achieves demand-sized, address-stable state allocation: it provisions contiguous storage sized strictly to demand for the entire lifecycle of the stream and implements retention through in-place writeback, eliminating memory reallocation and address drift.
To resolve the second question, \methodname realizes exact-shape replay without padding: it enumerates all reachable tensor shapes into a compact catalog before any inference traffic arrives, allowing hot operations to be captured and replayed at their exact geometries.
Crucially, these two rules are mutually enabling, and neither alone suffices (Table~\ref{tab:factorial}).
Shrinking memory without graph recording leaves the thousands of kernel launches untouched and yields no speedup.
Conversely, recording execution without demand-sized allocation forces the runtime to freeze state at the bloated envelope dictated by static constants, driving peak memory to its maximum at the exact granularity where recording is most effective.
Taken together, each rule supplies the precondition for the other: state reallocation shrinks the memory envelope that graph recording must freeze, while exact-shape recording enables demand-sized state to execute with zero launch overhead, delivering latency and memory benefits simultaneously at the same binding granularity (\S\ref{sec:formulation}).

We also subject the entire method to a non-negotiable prerequisite: the generated output must be identical \emph{byte for byte}.
Common acceleration techniques often pad inputs to predefined bucket capacities and execute masked operators~\cite{kwon2023vllm,paszke2019pytorch}, which can introduce numerical drift and produce differing output bytes.
For interactive products delivering audio directly into a listener's ear, altering waveforms modifies the delivered artifact itself, potentially inducing perceptual distortion or downstream transcription errors.
In our early explorations, an alternative decoding transformation improved p95 latency by 16--20\%, but it caused the generated token trajectory to diverge and raised character error rates; we therefore discarded it.
All optimizations in \methodname eliminate padding by enumerating exact shapes, treating byte-for-byte equality as an invariant checked before every call rather than an empirical property sampled post hoc.

This paper makes the following contributions:
\begin{itemize}
\item We identify and formalize the core impediment in intra-session real-time serving of full-duplex speech: the runtime lacks the model's \textbf{native clocks}.
Consequently, it can neither determine how much persistent state remains live nor anticipate reachable tensor shapes, forcing the system to over-provision memory and launch thousands of small operators sequentially.
We show that this workload class exhibits two statically verifiable properties: state liveness is bounded by retention policies (\emph{retention boundedness}), and timeline rates are fixed by configuration (\emph{clock declarability}).
These two properties are to intra-session real-time speech serving what \emph{batchability} is to LLM serving.

\item We design and implement \methodname, a runtime system structured around a declarative \textbf{native-clock contract}.
The contract exposes timeline rates and retention bounds, enabling two coordinated mechanisms: \emph{demand-sized, address-stable state allocation} and an \emph{exact-shape replay engine}.
Three key mechanisms make both rules hold simultaneously: decoupling stored extent from read extent to prevent memory shrinkage from conflicting with shape enumeration; enforcing in-place writeback to contract-owned memory to ensure pointer stability; and compiling exact reachable shapes without padding so that byte-exact equality is enforced before every invocation, including operators with per-call stochasticity (\S\ref{sec:formulation}, \S\ref{sec:design:replay}).

\item We evaluate \methodname across four released full-duplex models spanning three distinct decoder architectures, gating all results on \textbf{byte-for-byte equality}.
On the streaming speech synthesis module shared by MiniCPM-o~4.5 and Lychee-FD, \methodname achieves $2.85\times$ the speed of the stock runtime with $38.8\%$ lower peak memory (\S\ref{sec:eval:factorial}), confirming empirically that both rules are required.
On the live duplex serving path, mean SPEAK time for a single stream drops from 14\% above the one-second cadence to 2\% below it, eliminating interactive lag (\S\ref{sec:eval:e2e}).
Under multi-session concurrency, \methodname completes a four-session workload that exhausts GPU memory on the stock baseline within approximately 26\,GiB (\S\ref{sec:eval:e2e}, \S\ref{sec:appendix:capacity}).
\end{itemize}

The implications extend beyond speech.
Whenever an interactive session's stages execute sequentially and results must be delivered on a strict cadence, fixed launch overhead cannot be amortized across requests.
Real-time streaming translation, frame-by-frame video generation, and closed-loop robotic control all share this execution profile.
Because their internal states also advance at rates fixed by configuration, the native-clock contract offers a general principle to reclaim these system overheads.

\section{Background and Motivation}
\label{sec:motivation}

\subsection{Multi-Clock Real-Time Streaming Runtimes}
\label{sec:motivation:class}

We study the class of runtimes defined by three properties, each
checkable statically before any traffic arrives:

\begin{description}[leftmargin=*,itemsep=1pt,topsep=1.5pt,parsep=0pt]
\item[C1 (multiple native clocks).] Execution advances along two or more
  clocks whose rates are statically fixed by configuration across input
  lengths: LLM tokens, audio/codec frames, codebook depth, diffusion
  solver steps, waveform samples, and turns.
\item[C2 (bounded retention).] At least one persistent state region's
  extent obeys a \emph{declared retention policy}: a sliding window, a
  ring cache, or prompt plus retained history.
\item[C3 (single-session interactivity).] The runtime serves one
  interactive session whose stages are sequentially dependent, so
  intra-session batching is unavailable; cross-session batching remains
  possible and complementary.
\end{description}

Properties C1 and C2 represent the concrete, statically checkable form
of the two characteristics introduced in \S\ref{sec:introduction}: rates
fixed by configuration enable clocks to be explicitly declared, while a
declared retention policy strictly bounds state liveness.  Property C3
makes both characteristics operationally critical, because sequential
dependencies within an interactive session preclude intra-session batching.

MiniCPM-o~4.5~\cite{cui2026minicpmo45}, Moshi~\cite{defossez2024moshi},
and PersonaPlex~\cite{roy2026personaplex} satisfy all three conditions,
with C2 readable directly from a released configuration field.
Lychee-FD~\cite{liu2026lycheefd} and
Freeze-Omni~\cite{wang2025freezeomni} satisfy C2 through bounded
streaming caches whose extent is determined by an operator-chosen chunk
policy; they join the class once that policy is fixed, which is the form
our declaration takes (\S\ref{sec:appendix:membership}).  Offline TTS (lacking C3) and
unbounded LLM serving (lacking C2) fall outside this scope.

Property C3 distinguishes this workload class from the
conventional LLM-serving literature.  Continuous
batching~\cite{yu2022orca,kwon2023vllm}, chunked
prefill~\cite{agrawal2024sarathi}, and disaggregated
serving~\cite{zhong2024distserve} all amortize per-invocation launch
cost across independent requests, but that lever cannot be applied
\emph{inside} an individual session.  This sequential dependency entails
the corollary that defines this paper's second evaluation metric:
because invocations cannot be merged within a session, serving
multiple sessions on a single device requires holding their states
\emph{resident simultaneously}.  Consequently, peak memory allocation
directly governs system serving capacity.  Recent speech and
omni-modal serving systems~\cite{yin2026vllmomni,kamahori2026voxserve,
zhi2026liveserve} schedule stage invocations across independent
requests or sessions; in contrast, \methodname optimizes the
sequentially dependent execution path within an individual session.  These two perspectives operate at
distinct, complementary layers and compose naturally
(\S\ref{sec:related}; measured in Table~\ref{tab:phasepack}).

Satisfying these conditions does not by itself make an operator amenable
to exact specialization.  An auxiliary quantity,
derived from arithmetic on the same declared configuration fields,
determines whether the exact execution path is computationally
affordable, placing different stages of the same model on opposite sides
of the optimization boundary (\S\ref{sec:formulation:scope}).

\subsection{Anatomy of a Full-Duplex Second}
\label{sec:motivation:anatomy}

Execution profiling reveals two pronounced inefficiencies.  Each
corresponds to a theoretical lower bound implied by the model's published
configuration, yet both measured quantities sit far above their ideal
limits.

\subsubsection{Where the time goes}
\label{sec:motivation:time}

\begin{figure}[t]
\centering
\includegraphics[width=\columnwidth]{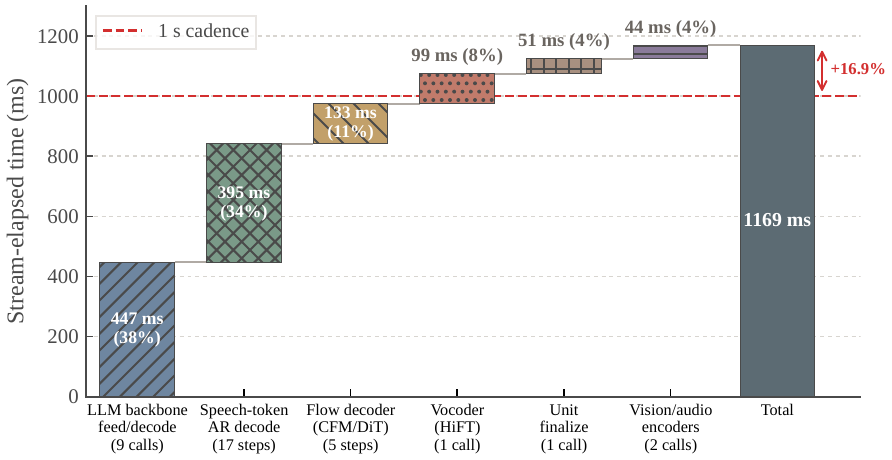}
\caption{\textbf{Stage time breakdown for one full-duplex second.}
MiniCPM-o~4.5 on A100 (omni-modal duplex, 17 emitted speech tokens; stream-elapsed times).
Autoregressive decoding fits within one second, whereas the synthesis tail causes a 16.9\% deadline overrun (\S\ref{sec:appendix:displaced}).}
\Description{Waterfall of stream-elapsed stage times for one SPEAK unit.
The LLM backbone and speech-token AR decode already fill one second;
the remaining stages push the total 16.9 percent over the 1 s cadence.}
\label{tab:speak_decomposition}
\end{figure}

Figure~\ref{tab:speak_decomposition} decomposes one representative SPEAK
unit of MiniCPM-o~4.5 on an NVIDIA A100 GPU (recorded using CUDA events;
emitting 17 speech tokens).  Across the profiled units, every second of
speech averages 9.5 backbone invocations (approximately 6.5 of them at
sequence length 1), followed by a five-step CFM solve with
CFG$\times2$, and one vocoder invocation.

The first two stages sum to $842$\,ms, comfortably fitting within the
one-second budget.  However, the subsequent synthesis tail requires an
additional $327$\,ms, pushing the total latency $16.9\%$ beyond the
cadence deadline.  The $2.85\times$ speedup reported in
\S\ref{sec:eval:factorial} is achieved specifically on this tail
segment, while the autoregressive fraction falls outside the lossless
scope for reasons analyzed in \S\ref{sec:formulation:scope}.

Comparing the hot operators against their compute lower bounds quantifies
this inefficiency.  Let an invocation of operator $o$ issue $m_o$
kernels with device execution times $t_k$ and a mean per-launch host
cost $\lambda$.  Let $G_o$ denote the \emph{orchestration slack} that a
launch trace attributes to host issue or synchronization, and let $L_o$
represent the roofline bound:
\begin{equation}
\label{eq:slack}
\begin{aligned}
L_o&=\max\big(\Phi_o/P,\ B_o/\beta_{\mathrm{dev}}\big),
\quad T_o=\textstyle\sum_{k\le m_o}t_k+G_o,\\
\gamma_o-1&=\frac{\sum_k t_k-L_o}{L_o}+\frac{G_o}{L_o},
\quad G_o\simeq m_o\lambda ,
\end{aligned}
\end{equation}
where $\Phi_o$ and $B_o$ denote the operator's FLOPs and memory traffic,
while $P$ and $\beta_{\mathrm{dev}}$ represent the device's peak compute
and bandwidth capacities.  The two terms in $\gamma_o-1$ capture
in-kernel inefficiency and orchestration overhead, respectively.
Because only the second term includes $G_o$, the slowdown factor
$\gamma_o$ can be bounded prior to full attribution: the flow, LLM, and
speech-token steps run at $31\times$, $\sim5\times$, and $>20\times$
their theoretical bounds (Table~\ref{tab:slack}).  The final relation
provides a first-order model that holds whenever host kernel issue does
not overlap with device execution, establishing launch count $m_o$
as the decisive lever for optimization.

\begin{table}[t]
\centering
\footnotesize
\caption{\textbf{Hot operators run far above their roofline bounds.}
Per-invocation time in ms against the theoretical roofline bound of \eqref{eq:slack} (\S\ref{sec:appendix:bounds}).}
\label{tab:slack}
\begin{tabular}{@{}lrrr@{}}
\toprule
Operator & Measured & Bound & $\gamma_o$\\
\midrule
Flow DiT step (CFG$\times$2) & 26.6 & 0.86 & 31$\times$\\
LLM step ($L{=}1$, 8B) & $\sim$50 & 10.3 & $\sim$5$\times$\\
TTS step ($L{=}1$, small) & $\sim$23 & $\ll$1 & $>$20$\times$\\
\bottomrule
\end{tabular}
\end{table}

\paragraph{Separating the two terms.}
Attributing $\gamma_o$ requires an experimental instrument that
eliminates $G_o$ while holding device execution time $\sum_k t_k$
strictly fixed.  Byte-exact replay provides precisely this capability:
by construction, a replayed invocation issues identical kernels over
identical tensor shapes in the exact same order, with SHA-256 hash
equality independently verifying output fidelity.  CUPTI profiling
across 20 matched chunks confirms that each kernel type maintains an
identical launch count in both execution arms.  Consequently, the delta
between eager execution and replay directly measures the host-induced
latency eliminated under an identical in-graph kernel sequence.  This
isolates $G_o$ from the eager baseline directly: eager device-busy time
accounts for only 103.7\,ms of a 241.2\,ms chunk, situating
$G_o\in[137,157]$\,ms, or 57--65\% of eager wall-clock time,
compared to less than 1\,ms of residual idle time following replay.
Orchestration slack thus constitutes the majority of
eager wall-clock time.  The two autoregressive rows lie outside the
lossless scope (\S\ref{sec:formulation:scope}); their $\gamma_o$ records
the same symptom arising from the same launch pattern.

This is the first of the two patterns that \methodname targets:
\emph{thousands of tiny, regularly shaped kernel launches per second
whose aggregate latency is dominated by launch and scheduling
overhead}.  Under Property C3, this entire overhead lands on
the session's critical path, and \eqref{eq:slack} names the operational
lever that remains.

\begin{table}[t]
\centering
\footnotesize
\caption{\textbf{Persistent state reserved envelopes versus declared live extents.}
Over-reservation ratios $\rho_r$ range from $3.5\times$ to $18\times$ across state regions.
Derivations are in \S\ref{sec:appendix:overreserve}.}
\label{tab:overreserve}
\begin{tabularx}{\columnwidth}{@{}>{\raggedright\arraybackslash}Xrrr@{}}
\toprule
Region and implementation constant & Reserved & Declared live & $\rho_r$\\
\midrule
Flow workspace (steps$\times$frames), duplex & 16$\times$1{,}000 & 5$\times$458 & \textbf{7.0$\times$}\\
\quad same, standalone decoder & 16$\times$1{,}000 & 10$\times$458 & \textbf{3.5$\times$}\\
Attention mask envelope (frames) & 500 & 56 & \textbf{8.9$\times$}\\
\quad largest bucket & 1{,}000 & 56 & \textbf{17.9$\times$}\\
Stacked carry per frame (fp32) & 2\,MiB & 0.125\,MiB & \textbf{16$\times$}\\
\bottomrule
\end{tabularx}
\end{table}

\subsubsection{Where the state goes}
\label{sec:motivation:state}

The same execution profile exposes the second inefficiency, which
is likewise measurable against a declared configuration bound.
Persistent state regions are provisioned according to static capacities
hardcoded in the implementation, whereas the declared retention policy
strictly bounds the extent that any downstream consumer ever accesses.
To formalize this disparity, we define the \emph{over-reservation
ratio} $\rho_r$, directly parallel to $\gamma_o$:
\begin{equation}
\label{eq:overreserve}
\rho_r=\frac{\text{reserved envelope}}{\text{declared live extent}},
\end{equation}
where the denominator, like $L_o$, is computed deterministically from
published configuration fields.

Table~\ref{tab:overreserve} evaluates $\rho_r$ across the state regions
examined in this paper, where the ratio ranges from $3.5\times$ to
$18\times$.  For example, the flow workspace alone reserves 2.1\,GB of
fp32 memory, whereas the declared live extent requires only one-seventh
of that allocation.  This over-provisioning represents the capacity
dimension of the problem.  Its second dimension, which is even more
disruptive for execution, concerns tensor \emph{position}.  Because
existing implementations apply retention policies via tensor
concatenation, they materialize a fresh multi-hundred-megabyte cache
tensor on every chunk.  This practice incurs redundant allocation
overhead, consumes memory bandwidth, and continuously moves
the state tensor to new memory addresses.  Consequently, any recorded
execution plan reused across chunks would immediately reference invalid
pointers.  The omission that prevents memory envelopes from shrinking to
true demand is precisely the one that prevents execution plans from being
frozen.  Across multi-session serving, these over-allocation
factors compound into session capacity limits (\S\ref{sec:eval:e2e}).

Together, these ratios quantify two coupled gaps stemming from the runtime's
inability to inspect native clocks: memory over-reservation
($\rho_r \in [3.5, 18]$) at drifting addresses, and orchestration slack
($\gamma_o \in [5, 31]$) under shapes the runtime treats as unknown.  Attempting to
remedy either gap in isolation worsens the other.

\subsection{Why Existing Tools Do Not Close the Gap}
\label{sec:motivation:whynot}

Existing optimization strategies fail for structural reasons predicted by
these coupled gaps (\S\ref{sec:eval:controls}).  First, the stock stack's
CUDA Graph branch uses hand-picked shape buckets that native clocks never
produce, so \emph{it never executes}; adjusting those buckets enables
execution but freezes padded envelopes at bucket capacity, preserving high
$\rho_r$.  Second, \code{torch.compile} in \code{reduce-overhead}
mode~\cite{paszke2019pytorch} aborts on key streaming paths and introduces
numerical drift while saving no memory, lacking visibility into state
liveness.  Third, graph recording in isolation shifts memory bloat to static
staging across shape classes, whereas state compaction alone leaves kernel
launch counts untouched.  The two inefficiencies cannot be resolved
independently, an empirically falsifiable proposition
(\S\ref{sec:formulation}).  Furthermore, lossy approximations compromise
fidelity: an early decoding transformation reduced p95 latency by 16--20\%
yet induced token divergence and non-zero text error rates, violating
conversational requirements.

Properties C1 and C2 jointly make this workload class tractable by
providing the structural information each gap lacks.  Because hot-operator
tensor shapes are deterministic functions of clocks with declared bounds,
all reachable shapes are known \emph{before inference traffic arrives}.
Concurrently, the declared retention policy defines live history extents,
making the denominator of \eqref{eq:overreserve} analytically computable at the
same moment.  Explicitly capturing native clocks enables runtimes to size
state to demand, anchor it to stable addresses, and replay execution plans
without padding.

\section{The Native-Clock Contract and Its Realization}
\label{sec:formulation}
\label{sec:design}

\begin{figure}[t]
\centering
\includegraphics[width=1\columnwidth]{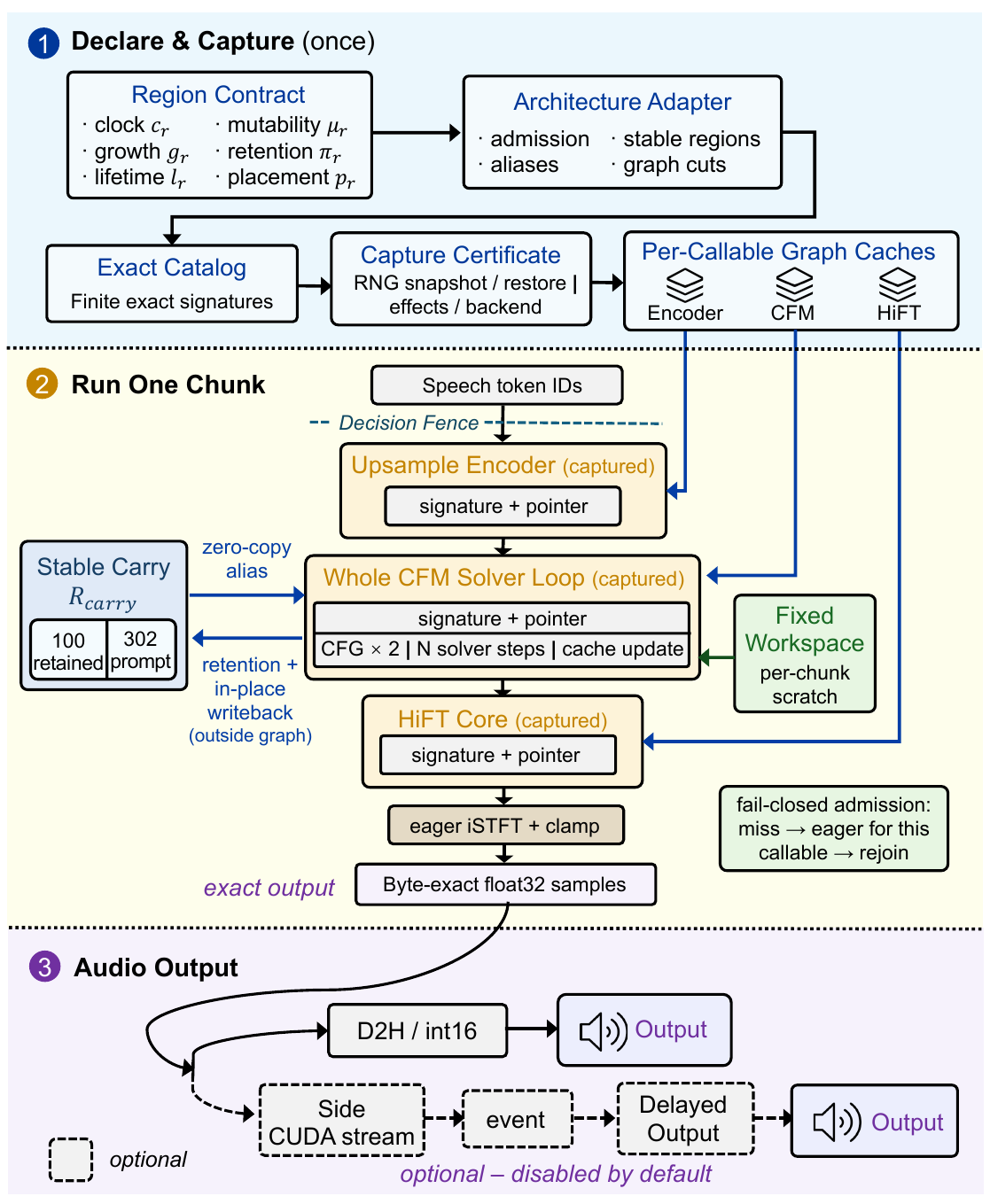}
\caption{\textbf{\methodname system architecture.}
Adapters declare native clocks, retention rules, and address-stable regions for demand-sized storage and a finite shape catalog.  At runtime, wrapped callables enforce fail-closed admission and write back in-place into the fixed-address carry.}
\Description{Three-stage DuplexCadence system diagram.  The declaration and
capture plane derives a demand-sized region layout and per-callable graph
caches.  The runtime plane applies a local fail-closed gate to the
upsampling encoder, the whole CFM solver loop, and the capturable
vocoder core.  Audio materialization follows, with an optional
side-stream path disabled by default.}
\label{fig:overview}
\end{figure}

Figure~\ref{fig:overview} depicts the architecture of \methodname: a
declarative contract instantiated by a per-family adapter, and two
coordinated rules derived from it.  Specifically, the adapter names
clocks, retention policies, and address-stable regions; the state rule
sizes these regions and pins the retained cache; and the compute rule
captures one CUDA graph per exact signature those clocks produce.
Omitting either rule leaves one performance axis unaddressed.  Below, we
state each rule in conjunction with the mechanism that realizes it.

\subsection{The Region Contract}
\label{sec:formulation:contract}
\label{sec:design:adapter}

We model a full-duplex runtime as an ordered sequence of operator
invocations reading immutable parameters, ephemeral workspace memory,
persistent state partitioned into regions $\{S_{u,r}^t\}$, and
pseudorandom generator states $\Xi^t$ (\S\ref{sec:appendix:invariant}).

\begin{definition}[Region contract]
\label{def:contract}
A state region $r$ is described by a tuple
$R_r=(\boldsymbol c_r,\mathbf g_r,\ell_r,\mu_r,\pi_r,p_r)$: its native
clock $\boldsymbol c_r$ (scalar or product, such as
$(\text{solver step},\text{mel frame})$); its unconstrained growth map
$\mathbf g_r$ over those clocks; its lifetime $\ell_r$ (operator, unit,
turn, session, process); its mutability class $\mu_r$; its retention policy
$\pi_r$, which bounds its extent as required by Property C2; and its
physical placement $p_r$.
\end{definition}

An adapter explicitly declares these structural properties, allowing the
runtime to derive all downstream execution parameters.  Per model family,
it specifies clock quantities, an admission predicate, and address-stable
regions, alongside the value-semantics and effect declarations detailed in
\S\ref{sec:appendix:engine}.  For the shared speech synthesis decoder at
our primary operating point, the complete declaration consists of:
$f_{\text{prompt}}{=}302$, $f_{\text{retained}}{=}100$,
$f_{\text{chunk}}{=}50$, $f_{\text{call}}{=}56$ frames,
$n_{\text{steps}}{=}10$, an upsampling rate of 2 frames per token, an
admission predicate $f\le f_{\text{chunk}}$, two address-stable regions,
and three device-resident constants (Table~\ref{tab:adapter}).  Every
subsequent extent and class count is derived deterministically from these
numbers.

From this declaration, the runtime establishes that a region's spatial
extent, and therefore its memory footprint, is governed entirely by its
clocks:
$\mathbf a_r(t)=\pi_r\big(\mathbf g_r(\boldsymbol c_r(t))\big)$, so that
$\bytes(S_{u,r}^{t})$ equals $\beta_r\prod_j a_{r,j}(t)$ at element width
$\beta_r$.  Property C2 ensures that $\pi_r$ caps each spatial coordinate
independently of session length, giving
$A_{r,j}=\sup_t a_{r,j}(t)<\infty$.  In contrast, stock implementations
provision memory at static constants $\hat A_{r,j}$ chosen without
regard to $\pi_r$, yielding the over-reservation ratio $\rho_r$ defined in
\eqref{eq:overreserve}.

\paragraph{Two extents, and therefore two regions.}
A single frame clock governs two distinct physical quantities that must be
kept apart.  The \emph{envelope extent} represents the maximum in-flight
capacity that physical storage must accommodate, because truncation under
the retention policy occurs strictly \emph{after} inference, meaning
newly presented frames must first fit within the buffer; this extent
governs byte allocation.  Conversely, the \emph{attended extent} denotes
the history span downstream consumers actually read, which specializes the
operator's tensor signature and fixes the number of reachable shape
classes.  While these two extents coincide under pure concatenation, they
diverge in the streaming flow decoder evaluated here, where lookahead
frames reside in the envelope but are discarded prior to truncation, thus
never specializing a signature.  This distinction produces two
contract-managed, address-stable regions: the \emph{carry}, sized to the
attended extent $L=f_{\text{prompt}}+f_{\text{retained}}$ (402 frames of
retained history propagated across chunks), and the \emph{solver
workspace}, sized to the envelope extent $10{\times}458$ frames (reducing
$\rho_r$ from $3.5\times$ standalone and $7.0\times$ in full-duplex vs.\ stock $16{\times}1000$).

Two additional clock classes become essential downstream, both expressible
under Definition~\ref{def:contract}.  A \emph{constant} has a trivial clock.
Consequently, rematerializing one per call, such as transferring a
host-side frequency table to the device at every diffusion step,
erroneously attributes it to the call clock, which \eqref{eq:slack} charges
as orchestration slack $G_o$.  The adapter remedies this by migrating
such constants to device-resident buffers with bitwise-identical contents
(\S\ref{sec:appendix:residency}).  Meanwhile, the \emph{device random
stream} advances along the call clock, as its Philox generator offset
advances by a deterministic increment per kernel invocation.

The losslessness of demand sizing is anchored in a representation
invariant beyond raw byte counts, formalizing \eqref{eq:representation}:
every declared consumer reads precisely what it would have obtained from the
logical region, while the region's memory pointer remains invariant across
its declared lifetime.  Because only the attended extent participates in
consumer computation, auxiliary capacity within the envelope remains free
of correctness hazards.

\begin{definition}[Strict trace equivalence]
\label{def:exact}
Let $F(X;\Xi)$ denote the externally observable execution trace of the
runtime on input $X$ and generator collection $\Xi$, encompassing
control decisions, emitted token identifiers, turn boundary events, and
raw PCM sample byte streams at native output precision (float32 in our
stack).  An optimized runtime achieves \emph{strict trace equivalence} if
$F_{\mathrm{opt}}(X;\Xi)=F_{\mathrm{base}}(X;\Xi)$ under bit-for-bit
comparison, without numerical tolerance.
\end{definition}

Every optimization in this paper satisfies Definition~\ref{def:exact},
verified through the paired evaluation protocol of \S\ref{sec:eval:setup}.

\subsection{State Rule: Demand Sizing and the Carry}
\label{sec:design:state}

The state rule allocates every contract-managed persistent or scratch
region strictly to the extent implied by its declared clocks, lifetime,
and retention policy, establishing $A^{\dagger}=A$.  The $16{\times}1000$
envelope collapses to $10{\times}458$, and immutable prompt states become
copy-on-write across conversational turns.  In isolation, this represents
a pure memory allocation discipline; its speed contribution is unlocked
only through synergy with the compute rule.

The central structural mechanism is the \emph{carry}: a contract-owned
buffer holding the retained attention cache at a single, invariant physical
address throughout the stream's lifetime.  While the stock implementation
applies retention via dynamic tensor concatenation
(\S\ref{sec:motivation:state}), \methodname writes retained data directly
\emph{into} a pre-allocated buffer $C$ of fixed length $L$.  Letting $H$
denote the live cache tensor at extent $\ell>L$, and using $\Vert$ for
concatenation along the frame clock dimension:
\begin{equation}
\label{eq:carry}
C[\,\cdot,0{:}L\,]
\;\longleftarrow\;
H[\,\cdot,0{:}f_{\text{retained}}\,]
\;\Vert\;
H[\,\cdot,\ell{-}f_{\text{prompt}}{:}\ell\,],
\end{equation}
producing elementwise-identical values under two fundamentally different
allocation disciplines.  The stock path evaluates the right-hand side by
allocating a new tensor on every chunk.  In contrast, \methodname writes
into existing storage, ensuring $\mathrm{ptr}(C)$ remains constant across
the stream lifecycle, including turn transitions where cache resets write
base state in-place into existing storage.  The slice indices
in \eqref{eq:carry} follow the cache layout of this model family, which
uses prepend ordering where the mutable history window precedes the
immutable prompt tail.  As reported in \S\ref{sec:design:verify}, this
layout invariant is formally verified at runtime.

The carry is a general memory discipline across architectures.  On
Moshi, it degenerates to the vendor's ring buffer; on Freeze-Omni, it
becomes a compact convolutional left-context buffer, both from the same
contract (\S\ref{sec:appendix:families}).

\subsection{Compute Rule: Exact-Shape Replay}
\label{sec:formulation:props}
\label{sec:design:replay}

The contract that bounds memory footprint simultaneously bounds the
universe of reachable tensor shapes, elevating the native clock from a
storage concept to an \emph{execution} control mechanism.  Let $\Sigma_o$
denote the mapping from admissible attended extents $\mathcal A_o$ of the
regions accessed by operator $o$, combined with its non-extent
configuration $\mathcal Q_o$ (data type, memory layout, device target,
and scalar arguments), to fully specialized tensor signatures.  Property
C2 ensures this signature space is finite independently of session
duration.  For the streaming flow decoder, the frame-clock component
forms a concise, closed arithmetic grid:
\begin{equation}
\label{eq:catalog}
\begin{aligned}
\catalog(o)&=\Sigma_o(\mathcal A_o\times\mathcal Q_o),
\quad
K=\big\lceil f_{\text{retained}}/f_{\text{chunk}}\big\rceil+1,\\
\mathcal L&=\{f_{\text{prompt}}+k\,f_{\text{chunk}}\}_{k=0}^{K-2}
\cup\{f_{\text{prompt}}+f_{\text{retained}}\}.
\end{aligned}
\end{equation}
Here, $\mathcal L$ collects the \emph{attended} extents of
\S\ref{sec:formulation:contract}.  For the Token2Wav module shared by
MiniCPM-o~4.5 and Lychee-FD (CosyVoice2 lineage~\cite{du2024cosyvoice2}),
$\mathcal L$ traverses the grid until truncation locks it at the retention
bound, yielding $\mathcal L=\{302,352,402\}$ and $K{=}3$ distinct shape
classes.  Moshi's ring cache gives $K{=}1$, and Freeze-Omni's streaming
policy gives $K{=}2$.  Because the stride follows the advancement clock
(distinct from the invocation frame count $f_{\text{call}}$), live calls
systematically fall outside the vendor's handcrafted bucket sets.  Partial
stream flushes can land between grid points, but the admission predicate
caps them by $f_{\text{chunk}}$, ensuring they remain within a finite,
pre-enumerable catalog.

The compute rule is realized by \code{ExactShapeReplay}, a general engine
that specializes a callable once per catalog signature over a CUDA Graph
backend~\cite{kwon2020nimble,paszke2019pytorch}.  Admission is governed by
predicate \eqref{eq:admit} evaluated over the captured subset
$\mathcal C_o\subseteq\catalog(o)$, failing closed at the first mismatch:
the first two conditions are checked per invocation at runtime, while the
third is certified at capture.

\textbf{(i) Signature match.}  The full tensor signature of the incoming
invocation (shapes, strides, data types, device, and scalar parameters)
must match an entry in $\mathcal C_o$.  Each specialized graph matches at
exact declared dimensions; any miss falls back safely to eager execution,
as does an argument the engine refuses to key on.

\textbf{(ii) Address stability.}  Regions written by the operator via
fixed addresses, declared by the adapter, are checked by data pointer,
accommodating callers that pass fresh views on identical storage.  If a
pointer drifts, the engine conservatively drops the class cache and logs
an invalidation.  Conversely, contract-owned input buffers like the carry
are \emph{aliased}: the engine records physical pointers once, bypasses
per-call staging copies, and validates pointer constancy at zero overhead.

\textbf{(iii) Effect classification.}  Prior to capturing a class, the
engine executes a probationary eager call to monitor host and device
random generator states, rewinding state immediately to keep the probe
transparent.  Invocations reading \emph{host} randomness would freeze
drawn values as constants within the graph and are permanently rejected.
In contrast, invocations reading \emph{device} randomness are admitted,
because the Philox generator offset advances deterministically along a
declared clock; furthermore, graph capture preserves RNG state neutrality
via snapshot and restoration, ensuring initial replay draws the exact same
sequence as eager execution.  This admits the HiFT vocoder, whose harmonic
generator samples device randomness on every call.  To our knowledge, no
prior replay system specializes randomness-drawing operators while
guaranteeing strict bit-for-bit equality.  Other external effects are
audited at capture boundaries and rejected if ambiguous
(\S\ref{sec:appendix:hazards}).

\begin{proposition}[Exact-shape replay]
\label{prop:replay}
Assume clauses (i)--(iii) hold, every non-aliased input is staged into its
captured buffer, and capture certification confirms deterministic kernel
dispatch, identical CUDA stream and event dependencies, and no unmodeled
host side effects.  Then graph replay reproduces the exact same state
transitions on outputs, persistent buffers, and generator states as eager
execution, bit for bit.
\end{proposition}

The premises of Proposition~\ref{prop:replay} are formally established in
\S\ref{sec:appendix:proofs}.  Their critical practical consequence is that
execution plans are captured at \emph{exact} geometric shapes.  In
contrast, bucketed or padded captures execute masked operator variants
with altered reduction trees, preserving eager fidelity only up to
floating-point reassociation error.

\paragraph{Granularity and capture placement.}
The shape catalog is an intrinsic property of the declaration, independent
of the clock level at which graph binding occurs.  Consider a chunk
enclosing $n_{\text{steps}}$ internal solver steps, where internal step
indices do not specialize the tensor signature and each step follows an
identical control path.  In this setting, binding at the chunk clock
preserves the total class count while reducing the host launch count by a
factor of $n_{\text{steps}}$.  Since orchestration slack scales with
kernel launches as $G_o\simeq m_o\lambda$ in \eqref{eq:slack},
this coarsening yields substantial latency benefits:
\begin{equation}
\label{eq:granularity}
|\catalog_{\text{chunk}}|=|\catalog_{\text{step}}|=K,
\qquad
\frac{m^{\text{host}}_{\text{chunk}}}{m^{\text{host}}_{\text{step}}}
=\frac{1}{n_{\text{steps}}}.
\end{equation}
At \emph{step} granularity, the engine captures individual DiT solver
steps, paying inter-step host coordination $n_{\text{steps}}$ times per
chunk.  At \emph{chunk} granularity, the engine encapsulates the whole
\texttt{forward\_chunk} into a unified CUDA graph, maintaining three
signature classes on either clock.  However, coarsening also raises memory
footprint: the chunk graph's input footprint includes the full stacked
carry, amounting to approximately 527\,MB in fp32 once demand-sized,
compared to 0.84\,GB under implementation-constant bounds
(\S\ref{sec:appendix:enablingderiv}).

\label{sec:design:lifecycle}%
Importantly, graph admission is evaluated \emph{locally}: each wrapped
callable makes independent decisions.  Consequently, an unexpected
signature miss incurs the cost of a single eager invocation while leaving
other cached graphs fully operational.  Furthermore, computing the lookup
key involves only integer and enum comparisons without inspecting tensor
data, ensuring admission checks impose zero overhead on the GPU\@.  In
steady state, all eager fallbacks are avoided, reducing an entire audio
chunk to exactly three efficient graph replays, one per wrapped
callable (Figure~\ref{fig:lifecycle}).

\subsection{Why the Two Rules Are Inseparable}
\label{sec:design:interlocks}

\begin{proposition}[Mutual enabling]
\label{prop:enabling}
Under the capture architecture of \methodname, every non-aliased input
tensor requires class-specific static storage, and every write-through
region requires a frozen, address-stable memory envelope.  Consequently,
demand sizing and zero-copy aliasing jointly minimize the memory
overhead of exact replay across both terms of \eqref{eq:enabling}.
Conversely, applying demand sizing in isolation leaves the kernel launch
count $m_o$ unchanged, leaving orchestration slack $G_o\simeq m_o\lambda$
unaffected and offering no latency reduction.  Attempting to reduce class
counts via input padding alters the underlying mathematical computation,
violating Proposition~\ref{prop:replay}.
\end{proposition}

Let $\mathcal I_{o,\kappa}$ denote tensor inputs of operator $o$ under
class $\kappa$, let $\mathcal A_{o,\kappa}\subseteq\mathcal I_{o,\kappa}$
represent those the engine can \emph{alias} at contract-owned stable
addresses, and let $\mathcal W_o$ denote frozen write-through regions.
When provisioning static staging buffers for each signature class, aggregate
reserved memory for a graph binding is expressed as:
\begin{equation}
\label{eq:enabling}
M^{\mathrm{reserve}}_o=
\sum_{\kappa\in\mathcal C_o}
\sum_{x\in\mathcal I_{o,\kappa}\setminus\mathcal A_{o,\kappa}}
\!\!\bytes(x)
\;+\;
\sum_{r\in\mathcal W_o}\beta_r\!\prod_{j\le d_r}\! A^{\dagger}_{r,j},
\end{equation}
decomposing into a class-specific input staging term and a frozen
write-through envelope sized to $A^{\dagger}_{r,j}$.

In concrete execution, this relationship manifests across three seams.
First, \emph{a single pointer check satisfies two distinct system
requirements}: the carry's fixed address validates pointer stability in
\eqref{eq:admit} while establishing the safety condition for zero-copy
aliasing.  A single \texttt{data\_ptr} check thus guarantees execution
exactness while eliminating staging copies.  Without the state rule, the
caller would materialize a new cache buffer per chunk, forcing the engine
to maintain dedicated per-class clones of the stacked carry, which explains
the sharp memory surge in replay-only configurations.  Second, \emph{one
rule directly bounds what the other must freeze}: the second term in
\eqref{eq:enabling} scales with $A^{\dagger}$, which the state rule
contracts from $(16,1000)$ down to $(10,458)$.  Third, \emph{memory
ownership eliminates signature classes}, acting directly on catalog
cardinality.  Because both staging memory and launch reductions scale with
binding granularity (\S\ref{sec:appendix:enablingderiv}),
Proposition~\ref{prop:enabling} forms an empirically falsifiable Pareto
claim: only coupled rules achieve optimal latency without paying the
replay-only memory penalty; isolated single rules fail in opposite directions
(\S\ref{sec:eval:factorial}, Figure~\ref{fig:pareto}).

When $N$ independent sessions share one accelerator, this accounting
governs system capacity.  Because \methodname time-multiplexes a
\emph{single} set of graph bindings across sessions, increasing $N$
replicates only each session's lightweight state snapshot, so the memory
reductions of \S\ref{sec:motivation:state} translate directly into
multi-session capacity (\S\ref{sec:appendix:sessions}).

\subsection{Scope: Catalog Width}
\label{sec:formulation:scope}

For graph specialization to remain practical, the reachable signature
catalog must be strictly \emph{compact}, because each captured class
incurs warmup time and dedicated staging memory under \eqref{eq:enabling}.
Let the attended extent specializing operator $o$'s signature span
$[A_r^{-},A_r]$, where $A_r$ is the retention bound and $A_r^{-}$ the
lower bound once the policy fires.  Let $s_r$ denote extent advancement
per invocation.  Traversing the retention window $W_r=A_r-A_r^{-}$ step
by step yields:
\begin{equation}
\label{eq:width}
K_r=\big\lceil W_r/s_r\big\rceil+1,
\qquad
|\catalog(o)|\le|\mathcal Q_o|\!\!\prod_{r\in\regions_o}\!\!K_r
\end{equation}
distinct tensor extents, defining $K_r$ as the region's \emph{catalog width}.

\begin{criterion}[Lossless scope]
\label{thm:scope}
Assume $\Sigma_o$ is injective with respect to the attended extent of
region $r$, and that this extent advances from $A_r^{-}$ by a fixed stride
$s_r$, undergoing single-mode truncation upon exceeding $A_r$.  Then the
reachable catalog cardinality is governed by \eqref{eq:width}, and an exact
ahead-of-time catalog requires at least $K_r$ specialized classes per
region.  Catalog width remains manageably small in precisely two regimes:
when the retention policy \emph{pins} the extent ($W_r=0$), or when the
invocation advances the extent in strides comparable to the window width
itself ($s_r=\Theta(W_r)$).  A policy that bounds the extent but advances
it by single units produces $K_r=W_r+1$.  Hence, bounded
retention under Property C2 is necessary but not sufficient for exact
specialization.  Compressing $K_r$ without coarsening the stride $s_r$
requires mapping multiple extents onto a single signature via padding,
which forfeits Proposition~\ref{prop:replay}.
\end{criterion}

Criterion~\ref{thm:scope} serves as an analytical tool evaluated directly
on configuration files prior to deployment.  As discussed in
\S\ref{sec:appendix:scope}, partial stream flushes at turn boundaries
temporarily relax the constant-stride condition, introducing at most $s_r$
additional classes derived from the same parameters.

\begin{table}[t]
\centering
\footnotesize
\caption{\textbf{Catalog width across declared retention policies.}
Evaluation of \eqref{eq:width} with $W_r$ and $s_r$ read from released configurations.
Regions with narrow catalog widths and stable signatures are captured, whereas large widths or variable query lengths fall outside the lossless scope.}
\label{tab:width}
\begin{tabularx}{\columnwidth}{@{}>{\raggedright\arraybackslash}Xrrrl@{}}
\toprule
Region and declared policy & $W_r$ & $s_r$ & $K_r$ & Verdict\\
\midrule
\multicolumn{5}{@{}l}{\emph{measured: catalogs captured in this paper}}\\
Moshi temporal KV, fixed-capacity ring & 0 & -- & \textbf{1} & captured\\
Token2Wav carry, 100 retained frames & 100 & 50 & \textbf{3} & captured\\
Freeze-Omni codec, chunk 40 $+$ pad 10 & 10 & 10 & \textbf{2} & captured\\
\addlinespace
\multicolumn{5}{@{}l}{\emph{computed from released configurations}}\\
Speech-token KV, windowed mode (steady) & 0 & -- & 1 & out: $\mathcal Q_o$\\
Language backbone, token watermark & 2{,}000 & 1 & 2{,}001 & out: width\\
\bottomrule
\end{tabularx}
\end{table}

Table~\ref{tab:width} illustrates both failure modes: the \emph{language
backbone} is excluded by catalog width ($K_r=2{,}001$ under unit-stride
decoding), while the \emph{speech-token decoder} satisfies spatial width
($K_r=1$) but introduces traffic-dependent query lengths into $\mathcal Q_o$.
Specialization reach thus depends jointly on retention configurations and
calling conventions, both computable from configurations prior to
implementation (\S\ref{sec:appendix:scope}).  \methodname's lossless
specialization focuses on narrow catalog widths and stable signatures,
cleanly composing with outer serving systems managing broader dynamic
regimes.

\subsection{What If the Declaration Is Wrong}
\label{sec:design:verify}

A declarative contract is only useful if the runtime can check its
premises, so \methodname treats \emph{verifiability} as part of the
contract.  \methodname discharges every premise of
Proposition~\ref{prop:replay} with a named mechanism that reports itself and
fails closed (Table~\ref{tab:certificate}).  The division of labor follows the
contract structure: premises that depend on a \emph{value} are re-tested every
call, since those are what traffic can break, while premises fixed by the
\emph{declaration} are settled once per class at capture, and a class that
fails certification never enters $\mathcal C_o$.  Every clause is counted, so
the gates of \S\ref{sec:eval:exact} test the certificate.

One mechanism audits the \emph{declaration} directly: a verification mode
evaluates the stock concatenation alongside every retention application and
compares the two elementwise.  Over four matched blocks the paths agree on all
264 retentions at a single address across 312 applications, turn boundaries
included, confirming \eqref{eq:carry} as a verified empirical invariant
across every call (\S\ref{sec:appendix:carry}).

During development, an initial append-ordered carry implementation matched
baseline values for two chunks before cache eviction reached the prompt
tail, diverging on chunk three.  The shadow auditor immediately flagged the
discrepancy and localized the responsible retention routine, confirming
that runtime auditing exposes latent violations even under
plausible audio.

\paragraph{Implementation.}
\label{sec:impl}%
\methodname is implemented in Python on an unmodified PyTorch~2.8.0 /
CUDA~12.8 stack.  The core engine, \code{ExactShapeReplay}, is 263
executable lines that name no model, decoder, or operator; an adapter
adds 126--314 lines of declarations and binding sites
(Table~\ref{tab:crossarch}).  Capture happens once and its cost is reported:
the segment's nine classes, the three declared extents of \eqref{eq:catalog}
on each of three captured callables, cost 4.9--8.8\,s of warmup, none of it
inside a measurement window.  Admission also fails closed, so a rejection
costs launch latency and never exactness, every rejection counted by reason.
The engine interface, the porting cost, the capture-storage accounting, the
determinism envelope, and the four PyTorch hazards the contract absorbs are
in \S\ref{sec:appendix:engine} and~\S\ref{sec:appendix:determinism}; the
artifact will be released.

\paragraph{An optional third rule.}
An optional \emph{schedule rule} reorders the synthesis segment onto a
side stream.  Implemented but shipped disabled, its paired
delta includes zero, and no result here depends on it
(Table~\ref{tab:o}, \S\ref{sec:appendix:orule}).

\section{Evaluation}
\label{sec:eval}

We organize our evaluation around four questions of progressively widening
scope, one per subsection: whether the state and compute rules are mutually
enabling (Proposition~\ref{prop:enabling}); whether segment-level gains
survive end-to-end integration into the live duplex path across latency,
memory, and bitwise exactness, and whether they expand concurrent session
capacity; whether the native-clock principle generalizes across disparate
neural architectures; and how \methodname compares against compiler and
vendor graph baselines.

Numerical exactness is enforced as a strict gating condition in every
table, with bit-for-bit equality guaranteeing perceptual audio quality.

\subsection{Setup}
\label{sec:eval:setup}

\paragraph{Hardware and workloads.}
Primary results run on an NVIDIA A100 (40\,GB SXM4), with the factorial,
the serving points, and end-to-end MiniCPM-o re-run on an L20 (48\,GB) to
assess cross-generational behavior (\S\ref{sec:eval:hardware}).  We
evaluate four distinct workloads: (1) the \emph{shared decoder},
comprising the streaming Token2Wav pipeline (UpsampleConformer $\to$
CFM/DiT $\to$ HiFT) that MiniCPM-o~4.5 and Lychee-FD both deploy,
streaming 25-token chunks under the declaration of
\S\ref{sec:design:adapter}; (2) \emph{end-to-end} MiniCPM-o~4.5 in
omni-modal streaming mode under the two protocols Table~\ref{tab:e2e}
names; (3) \emph{cross-architecture} Moshi and Freeze-Omni's bounded
TiCodec generator, neither of which shares code with the first; and (4)
\emph{shared-weight multi-session}, where $N$ sessions time-multiplex one
process-scoped binding under a paced 2\,s-SLO harness.  Block counts,
seeds, chunk budgets, and pinned kernel settings are detailed in
\S\ref{sec:appendix:setup}.

\paragraph{Protocol and exactness ledger.}
\label{sec:eval:exact}
Every comparison uses \emph{matched blocks}: within a block, all variants
see the same seed, the same input stream, and a fresh process, and
reported values are per-block medians aggregated across blocks with
paired 95\% intervals inline.  Peak allocation is the caching allocator's
peak for the whole process, so every resident replay binding, staging
buffers and frozen envelopes alike, is inside it.  Cells are labeled
\textbf{B0} (baseline), \textbf{S} (state rule), \textbf{R}/\textbf{Rc}
(replay at step/chunk clock), and \textbf{SR}/\textbf{SRc} (both).  B0 is the stack as it ships \emph{and as it runs}: the vendor's recorded
fast path is present in code but never fires (\S\ref{sec:introduction}),
making eager issue the actual deployed baseline.  The repaired-bucket
variant in \S\ref{sec:eval:controls} serves as an auxiliary control
constructed specifically for comparison.  A cell that captures inside the
measured window is excluded from segment latency claims; in the live duplex
path that cost is charged to \methodname.

Exactness gates are pre-registered before measurement
(\S\ref{sec:appendix:protocol}) and \emph{every} \methodname performance
cell supporting a claim is gated against its unmodified baseline:
segment experiments hash float32 PCM streams; whole-system
experiments also compare token-ID sequences and discrete control
fields.  Non-exact controls are labeled.  The ledger also reaches
inside the run, comparing the representation invariant per call against
the concatenation it replaces (\S\ref{sec:design:verify},
\S\ref{sec:appendix:carry}).  Absolute times differ across harnesses and
kernel protocols, so every comparison below is drawn \emph{within} one
sweep (\S\ref{sec:appendix:provenance}) under stock kernels, the
protocol least favorable to \methodname (Table~\ref{tab:protocol}).  \label{sec:eval:micro}%
Admitting randomness-consuming operators presupposes that replay draws
the values eager execution would, which we verified
(\S\ref{sec:appendix:rng}); that clause admits the HiFT vocoder, worth
15.0\,ms per chunk.

\subsection{Evidence for Mutual Enabling}
\label{sec:eval:factorial}

\begin{table}[t]
\centering
\footnotesize
\caption{\textbf{Factorial evaluation of state and compute rules.}
Per-chunk p50 latency, vocoder time, and peak memory on A100 ($n{=}4$ matched blocks).
Speedup and $\Delta$Peak are relative to B0.  All 20/20 optimized cells achieve bit-exact PCM.}
\label{tab:factorial}
\begin{tabular}{@{}lrrrrr@{}}
\toprule
Variant & p50 & Voc. & Peak & Speedup & $\Delta$Peak\\
\midrule
B0 & 241.7 & 19.8 & 4,852 & --- & ---\\
S & 244.2 & 19.9 & 2,826 & 0.99$\times$ & $-$41.8\%\\
R (step) & 86.3 & 4.8 & 5,194 & 2.80$\times$ & $+$7.0\%\\
SR (step) & 86.3 & 4.7 & 3,109 & 2.80$\times$ & $-$35.9\%\\
Rc (chunk) & 86.6 & 4.9 & 7,943 & 2.79$\times$ & $+$63.7\%\\
\textbf{SRc (chunk)} & \textbf{84.9} & \textbf{4.7} & \textbf{2,972} & \textbf{2.85$\times$} & \textbf{$-$38.8\%}\\
\bottomrule
\end{tabular}
\end{table}

Table~\ref{tab:factorial} isolates the enabling structure.  Its
single-rule rows S, R, and Rc are the pre-registered \emph{falsifiers}
of Proposition~\ref{prop:enabling}.

In an eight-block evaluation where turn-boundary resets restore into the
carry as deployed in production, the CFM loop achieves 26/26 replays with
zero eager fallbacks, reproducing $-64.8\%$ latency (paired 95\% CI:
$[-65.3,-64.6]$) and $-38.5\%$ peak memory across 40/40 bit-exact blocks
(\S\ref{sec:appendix:admission} details the 24/26 harness-reset artifact in
Table~\ref{tab:factorial}).  Tail latency remains tightly bounded: SRc's
p95 is 85.3\,ms (p50: 84.8\,ms), confirming that median gains hold at the
tail.  Both single-rule falsifiers reproduce on a second GPU
(\S\ref{sec:eval:hardware}).

\emph{The state rule alone provides no reliable speedup}: its eight-block
point estimate of $-0.5\%$ carries an interval that crosses zero, so
demand sizing is a memory mechanism.  \emph{Replay alone costs memory}, and
the cost scales with granularity, because at the chunk clock the engine is
stuck with the implementation-constant envelope \emph{and} must clone the
stacked carry per captured class.  It also captures four classes (expanding
from three) due to the duplicate $\mathcal Q_o$ entry predicted in
\S\ref{sec:formulation:props} when no contract-owned carry collapses the two
strides (\S\ref{sec:appendix:admission}).  Equation~\eqref{eq:enabling}
prices the whole penalty from the declaration, predicting both peak deltas
to within a granule-independent residual (\S\ref{sec:appendix:enabling}).  \emph{Coupled, the two costs
cancel and the two gains survive}: no cell that omits a rule achieves both,
and both omissions fail in opposite directions, matching the proposition.

\paragraph{Coarsening polarity depends on state management.}
The coupling determines the direction of granularity scaling.
With the state rule active, coarsening from the step clock to the chunk clock
improves latency (p50: 86.3 to 84.9\,ms) and reduces peak memory (3,109 to 2,972\,MiB)
while reducing replay invocations tenfold (SR $\to$ SRc, Table~\ref{tab:factorial}).
Without the state rule, coarsening degrades both axes ($+56.7$\,pp peak memory;
$86.3\to86.6$\,ms, R $\to$ Rc) because un-aliased inputs must be re-staged on
every invocation, turning memory bloat into bandwidth bottlenecks on the critical
path.  The coarsened granularity providing maximum speedup is viable only
in conjunction with demand-sized state.

Decomposing the $-156.4$\,ms overall latency reduction attributes 135.7\,ms to
collapsing the solver loop, 15.0\,ms to admitting the stochastic vocoder via
effect classification, and 5.8\,ms to encoder and host glue
(\S\ref{sec:appendix:attribution}).  This recovers essentially all measured
orchestration slack ($G_o \in [137, 157]$\,ms in \S\ref{sec:motivation:time}),
leaving residual latency dominated by physical device kernels within the
graph (\S\ref{sec:appendix:attribution}).

\subsection{Surviving the Full Duplex Path}
\label{sec:eval:e2e}

\begin{table}[t]
\centering
\footnotesize
\caption{\textbf{End-to-end full-duplex evaluation on MiniCPM-o~4.5.}
A100, $n{=}6$ matched blocks.  Under the stock protocol (lower block, 48 units), mean SPEAK time drops below the one-second cadence (\S\ref{sec:appendix:provenance}).}
\label{tab:e2e}
\begin{tabularx}{\columnwidth}{@{}>{\raggedright\arraybackslash}Xrrr@{}}
\toprule
Metric & B0 & Rc & \textbf{SRc}\\
\midrule
\multicolumn{4}{@{}l}{\emph{deterministic protocol}}\\
Synthesis segment (ms) & 201.7 & 72.4 & \textbf{72.2}\\
Peak allocation (MiB) & 23,545 & 26,746 & \textbf{21,198}\\
Token and PCM SHA & --- & exact & exact\\
\addlinespace
\multicolumn{4}{@{}l}{\emph{stock protocol (cadence)}}\\
SPEAK mean (ms) & 1,141 & --- & \textbf{976}\\
\quad vs.\ 1\,s cadence & $+$14.1\% & --- & \textbf{$-$2.4\%}\\
SPEAK p50 (ms) & 1,169 & --- & \textbf{994}\\
SPEAK p95 (ms) & 1,312 & --- & \textbf{1,134}\\
\bottomrule
\end{tabularx}
\end{table}

Inside the live model the decoder shares a device and an allocator with
an 8B backbone, a vision encoder, and a streaming duplex loop, and
Table~\ref{tab:e2e} meets the three criteria we fixed in advance for
that setting.  The speedup transfers: the synthesis segment falls
from 201.7 to 72.2\,ms, reproducing the standalone $2.85\times$ to within
3\%.  Exactness holds over the whole system: token-ID sequences
and float32 PCM traces of all six matched blocks match the unmodified
baseline exactly, with control decisions and turn boundaries matching as
discrete fields.  The memory saving also compounds: process peak
drops by 10\%, whereas replay without the state rule \emph{raises} it by
13.6\%, reproducing the mutual-enabling signature at whole-system scale,
where the penalty is measured in gigabytes.  The two arms are level on
the segment they both transform, so they differ by 5,548\,MiB, which a
latency-only deployment would give up without gaining any speedup.

Admission in the live path matches the share predicted in
\S\ref{sec:design:replay}: the CFM loop replays 25 of 31 invocations, the six misses
being the cache-free prompt-priming pass, sent to eager by construction.
One declared class per block is captured inside the window and charged
to \methodname (\S\ref{sec:appendix:admission}).

The lower block measures cadence directly under the stock kernel
protocol, with no cross-harness substitution.  Mean SPEAK time falls
from 1{,}141 to 976\,ms: a single stream goes from 14.1\% over the
one-second period at which its input arrives to 2.4\% under it, which is
to say that on average it keeps up with the person speaking.  The p50 is
0.6\% under as well and the p95 remains 13\% over, so the claim is the
typical unit, not every unit; the remaining overrun comes from the
autoregressive stages that \S\ref{sec:formulation:scope} places outside
the lossless scope.

\begin{figure}[t]
\centering
\includegraphics[width=\columnwidth]{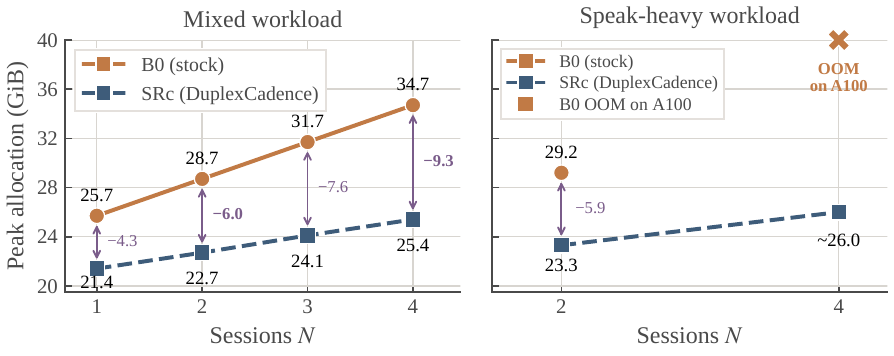}
\caption{\textbf{Serving capacity under shared-weight multi-session concurrency.}
Peak memory allocation in GiB on A100.  \methodname consistently reduces peak allocation across session counts, enabling the speak-heavy $N{=}4$ workload to complete within $\approx$26\,GiB where baseline runs out of memory (Table~\ref{tab:capacity}).}
\Description{Two-panel line plot of peak GPU allocation versus session
count.  Mixed and speak-heavy workloads both show a widening gap
between the stock baseline and DuplexCadence; speak-heavy N=4 is OOM on
A100 for the baseline.}
\label{tab:servingcap}
\end{figure}

\paragraph{Serving capacity.}
Under C3 the memory a session returns is what lets another session onto
the device, which we test on a shared-weight harness where $N$ sessions
time-multiplex one process-scoped binding.  The quantitative axis is
peak allocation and feasibility, not SLO-goodput. Peak allocation falls
in 24 of 24 matched pairs (Figure~\ref{tab:servingcap}), by
6.0\,GiB at mixed $N{=}2$ and 9.3\,GiB at $N{=}4$, because the catalog
and the carry are process-scoped and shared.  The practical consequence is
a change in feasibility: under a speak-heavy $N{=}4$ trace the baseline
exhausts device memory in all six blocks and produces no measurement at
all, while \methodname completes every block within $\approx$26\,GiB.  At
$N{=}1$ both arms sit on the harness ceiling of $0.5$, so that cell does
not read out the latency gain.  Paired PCM byte equality is 100\%
throughout, and SLO-goodput moves the same way with confidence intervals
too wide to support a claim (\S\ref{sec:appendix:capacity}).

\subsection{Cross-Architecture Evidence}
\label{sec:eval:crossarch}

\begin{table}[t]
\centering
\footnotesize
\caption{\textbf{Cross-architecture generalization across disjoint decoders.}
Medians of paired speedup ratios against eager execution; Exact counts gated blocks with bit-identical outputs; LOC denotes executable adapter lines (\S\ref{sec:appendix:porting}).}
\label{tab:crossarch}
\begin{tabularx}{\columnwidth}{@{}>{\raggedright\arraybackslash}Xrrrr@{}}
\toprule
Stack and mode & Latency & Speedup & Exact & LOC\\
\midrule
Token2Wav, exact catalog (SRc) & 84.9\,ms/chunk & 2.85$\times$ & 20/20 & 314\\
Moshi, vendor graph (width-1 check) & 31.6\,ms/frame & 4.97$\times$ & yes & 0\\
Moshi, \texttt{ExactShapeReplay} & 44.3\,ms/frame & 3.46$\times$ & 3/3 & 133\\
Freeze-Omni, TiCodec generator & 5.0 vs.\ 8.6\,ms & 1.72$\times$ & 3/3 & 126\\
\bottomrule
\end{tabularx}
\end{table}

On Moshi, the vendor implements custom temporal ring buffers and
single-graph wrappers (catalog width one in \eqref{eq:width};
\S\ref{sec:appendix:moshi}).  Toggling this vendor replay yields
$4.97\times$ speedup per frame to satisfy Moshi's 80\,ms real-time
budget.  Peak allocation nonetheless \emph{falls} by 652\,MiB
($-$3.8\%) when the toggle goes on, even though the wrappers alias
nothing and re-copy every tensor argument on every call --- exactly the
$\mathcal I_{o,\kappa}\setminus\mathcal A_{o,\kappa}$ staging term of
\eqref{eq:enabling}, which places them at the Rc cell of
Table~\ref{tab:factorial}.  The ring has already right-sized everything
there is to clone, so replay over declaration-bounded state is nearly
free in memory, and the $+63.7\%$ Rc pays in
Table~\ref{tab:factorial} is the price of replaying over state that no
declaration bounds --- Proposition~\ref{prop:enabling} confirmed from
the opposite direction.  Replacing these specialized wrappers
with generic \texttt{ExactShapeReplay} driven solely by declarative
ring-cache and frame-clock specifications recovers $3.46\times$ with 3/3
bitwise equality, capturing 89\% of the custom optimization's reduction
under a general, verifiable contract (\S\ref{sec:appendix:moshi},
\S\ref{sec:appendix:freezeomni}).

\subsection{Compiler and Vendor-Graph Controls}
\label{sec:eval:controls}

\begin{table}[t]
\centering
\footnotesize
\caption{\textbf{Comparison against compiler and vendor-graph controls.}
NVIDIA A100, $n{=}4$ matched blocks.  Latency p50 in ms, peak memory in MiB, and gap relative to \methodname SRc (positive means slower).  No control achieves speedup, memory reduction, and bitwise exactness together.}
\label{tab:controls}
\begin{tabularx}{\columnwidth}{@{}>{\raggedright\arraybackslash}Xrrrl@{}}
\toprule
Configuration & p50 & Peak & vs.\ ours & SHA\\
\midrule
\methodname SRc & \textbf{84.5} & \textbf{2,983} & --- & \textbf{equal}\\
B0 (eager) & 245.3 & 4,852 & $+196.8\%$ & ref.\\
\addlinespace
\multicolumn{5}{@{}l}{\emph{\texttt{torch.compile}, \texttt{reduce-overhead} mode}}\\
\quad DiT, \texttt{dynamic=False} & 122.3 & 4,852 & $+46.2\%$ & \textbf{differs}\\
\quad DiT, dynamic default & 123.1 & 5,048 & $+46.5\%$ & \textbf{differs}\\
\quad DiT, \texttt{max-autotune}$^{a}$ & 167.0 & 5,259 & $+102.3\%$ & \textbf{differs}\\
\quad DiT ${+}$ our capture & 117.2 & 5,077 & $+38.7\%$ & \textbf{differs}\\
\quad encoder / whole loop$^{b}$ & --- & --- & \multicolumn{2}{l}{does not run}\\
\addlinespace
Vendor graph, re-bucketed & 119.7 & 5,163 & $+43.3\%$ & equal$^{c}$\\
\bottomrule
\end{tabularx}

\raggedright\footnotesize Every paired 95\% CI on the gap column excludes
zero by a wide margin (\S\ref{sec:appendix:controlci}).
$^{a}$\texttt{max-autotune-no-cudagraphs}.  $^{b}$Three scopes, counting
\texttt{fullgraph=True}, each failing in 4/4 blocks in the two distinct
ways \S\ref{sec:appendix:hazards} details.  $^{c}$An empirical accident
of this kernel/shape combination; see text.
\end{table}

Across all evaluated configurations of \texttt{torch.compile} in \texttt{reduce-overhead} mode (Table~\ref{tab:controls}), the diffusion loop aborts inside Inductor, while the streaming encoder compiles but corrupts runtime graph outputs by violating stream-cache value semantics.  Enabling \texttt{max-autotune} increases latency (167.0\,ms), confirming that kernel compute efficiency is not the limiting factor.  Running compiler-generated kernels inside our exact-shape capture engine eliminates launch overhead yet remains 38.7\% slower with divergent bytes, so the numerical drift arises from Inductor's code generation.

Re-bucketing the vendor's graph branch to live chunk sizes achieves $-$52\% latency relative to eager execution, yet remains 43.3\% slower and 73\% larger than \methodname SRc due to padded envelope freezing.  Although its PCM matches eager execution bitwise under this specific kernel, input padding generally alters reduction trees and forfeits exactness guarantees while incurring dual-axis overheads.

\paragraph{Transfer to a second GPU\@.}
\label{sec:eval:hardware}
All findings generalize across GPU generations: on an L20, per-chunk speedup
reaches $2.95\times$ ($2.85\times$ on A100) alongside a $38\%$ peak memory
reduction, with both single-rule falsifiers reproducing.  The L20's 48\,GB
memory also measures the speak-heavy $N{=}4$ baseline (peaking at 40.2\,GiB),
confirming that A100 baseline failures reflect genuine memory exhaustion
(\S\ref{sec:appendix:l20}).

\subsection{Limitations}
\label{sec:eval:limits}

\textbf{Lossless scope.}  Exact specialization is bounded by catalog
width, leaving unit-stride autoregressive phases to outer serving
schedulers (Table~\ref{tab:width}).
\textbf{Harness.}  The shared-weight harness evaluates concurrency under
a 2\,s window, producing wide goodput intervals.
\textbf{Cross-session composition.}  Coordinating continuously advancing
sessions remains an open challenge addressed via process-scoped
time-multiplexing; prior systems~\cite{yin2026vllmomni,kamahori2026voxserve,
zhi2026liveserve} batch independent requests.
\textbf{Coverage.}  The cache-free prompt-priming pass and turn-boundary
pointer misses execute eager.
\textbf{Generality.}  Extending declarative contracts to LLM KV caches is
constrained by catalog width under bounded retention.
\textbf{Determinism envelope.}  Cross-process reproducibility inherits the
deterministic kernel library; unpinned eager baselines drift
(\S\ref{sec:appendix:limits}).

\section{Related Work}
\label{sec:related}

\methodname bridges model serving, deterministic execution, and streaming
state retention.  Where conventional runtimes pad inputs and batch
requests, \methodname derives demand-sized memory and exact-shape
execution from declarative native clocks, gaining speed and memory
jointly while analytically bounding specialization reach.

\paragraph{Serving systems.}
Cross-request batching, paged memory management, prefix caching, chunked
prefill, and disaggregated serving~\cite{yu2022orca,kwon2023vllm,zheng2024sglang,
agrawal2024sarathi,zhong2024distserve} amortize launch overhead across
requests.  Recent speech and omni-modal serving
systems~\cite{yin2026vllmomni,kamahori2026voxserve,zhi2026liveserve} extend
these techniques to audio via fixed chunking or bucketing.  These methods
operate across independent requests, orthogonally to our intra-session
optimization (\S\ref{sec:eval:limits}).  Nova~\cite{xu2025nova} pipelines
stages within a request to mitigate inter-stage latency, whereas
\methodname eliminates orchestration slack within individual stages.

\paragraph{Predictable execution.}
Clockwork~\cite{gujarati2020clockwork} schedules inference timing proactively, while Nimble~\cite{kwon2020nimble} and Rammer~\cite{ma2020rammer} reorder GPU execution to eliminate kernel launch overhead ($G_o$ in \eqref{eq:slack}).  Those systems infer timing from profiles or dataflow graphs for latency SLOs; in contrast, \methodname relies on declarative configuration timing to guarantee strict bitwise equality (Proposition~\ref{prop:replay}).

\paragraph{Shape specialization.}
Dynamic compilers~\cite{zhu2021disc,fegade2021cortex},
\texttt{torch.compile} in \texttt{reduce-overhead} mode~\cite{paszke2019pytorch},
and bucketed decode graphs~\cite{kwon2023vllm} handle shape variation by
padding or polymorphic dispatch.  \methodname enumerates all reachable
shapes exactly, avoiding padding masks and preserving numerical reductions;
Criterion~\ref{thm:scope} determines tractability ahead of time, and kernel
optimizations~\cite{dao2022flashattention} stay orthogonal.

\paragraph{Bounded retention for streaming state.}
Windowed attention, eviction policies~\cite{xiao2024streamingllm,
zhang2023h2o,li2024snapkv}, paged KV blocks~\cite{kwon2023vllm}, and ring
buffers bound unbounded context.  \methodname derives retention bounds from
configuration before allocation and applies them without eviction,
enabling exact-shape replay.

\paragraph{Approximation techniques and speech models.}
Quantization~\cite{xiao2023smoothquant}, fast ODE
solvers~\cite{lu2022dpmsolver,song2023consistency}, and speculative
decoding~\cite{leviathan2023speculative} modify computation, sacrificing
bitwise equality and risking conversational fidelity
(\S\ref{sec:motivation:whynot}).  Full-duplex speech
models~\cite{cui2026minicpmo45,defossez2024moshi,liu2026lycheefd,
wang2025freezeomni,roy2026personaplex,xu2025qwen25omni,wu2025stepaudio2}
and their synthesis pipelines~\cite{du2024cosyvoice,du2024cosyvoice2,
lipman2023flowmatching,tong2024conditionalflowmatching,peebles2023dit,
li2023hiftnet} optimize model representation and perceptual quality;
\methodname treats their temporal rates as an execution contract, unlocking
systems performance at zero quality loss.

\section{Conclusion}
\label{sec:conclusion}

Full-duplex speech models suffer latency and memory overheads because
runtimes cannot inspect native clocks governing state advancement.
\methodname elevates these timelines into a declarative contract, enabling
demand-sized state allocation at a stable address and exact-shape graph
replay.  Coupled on production decoders, the two rules
achieve $2.85\times$ speedup and $38.8\%$ lower peak memory with
bit-identical outputs.  On the live duplex path, mean SPEAK latency falls
from 14\% above cadence to 2\% below it, eliminating lag and expanding
multi-session capacity.  With optimization scope computable ahead of time, we propose retention
boundedness and clock declarability as principles for real-time
serving, parallel to batchability in LLM serving.

\bibliographystyle{ACM-Reference-Format}
\bibliography{refs}

\end{document}